# Inductive Kernel Low-rank Decomposition with Priors: A Generalized Nyström Method

**Kai Zhang**                                                                    kai-zhang@siemens.com
Siemens Corporate Research and Technology, Princeton, NJ

**Liang Lan**                                                                    lanliang@temple.edu
Department of Computer and Information Sciences, Temple University, Philadelphia, PA

**Jun Liu**                                                                      jun-liu@siemens.com
Siemens Corporate Research and Technology, Princeton, NJ

**Andreas Rauber**                                                               rauber@ifs.tuwien.ac.at
Institute of Software Technology and Interactive Systems, Vienna University of Technology, Vienna

**Fabian Moerchen**                                                              fabian.moerchen@siemens.com
Siemens Corporate Research and Technology, Princeton, NJ

## Abstract

Low-rank matrix decomposition has gained great popularity recently in scaling up kernel methods to large amounts of data. However, some limitations could prevent them from working effectively in certain domains. For example, many existing approaches are intrinsically unsupervised, which does not incorporate side information (e.g., class labels) to produce task specific decompositions; also, they typically work "transductively", i.e., the factorization does not generalize to new samples, so the complete factorization needs to be recomputed when new samples become available. To solve these problems, in this paper we propose an "inductive"-flavored method for low-rank kernel decomposition with priors. We achieve this by generalizing the Nyström method in a novel way. On the one hand, our approach employs a highly flexible, nonparametric structure that allows us to generalize the low-rank factors to arbitrarily new samples; on the other hand, it has linear time and space complexities, which can be orders of magnitudes faster than existing approaches and renders great efficiency in learning a low-rank kernel decomposition. Empirical results demonstrate the efficacy and efficiency of the proposed method.

## 1. Introduction

Low-rankness is an important structure widely exploited in machine learning. For example, the kernel matrix often has a rapidly decaying spectrum and thus small rank, and eigenvectors corresponding to large eigenvalues create useful basis function for classification (Williams & Seeger, 2000; Bach & Jordan, 2005). Therefore the low-rank constraint has been widely applied to kernel learning problems (Kulis et al., 2009; Lanckriet et al., 2004; Shalit et al., 2010; Machart et al., 2011). On the other hand, the low-rank property is very useful in reducing the memory and computational cost in large scale problems, most notably by the so called low-rank matrix decomposition. Such a decomposition produces a compact representation of large matrices, which is the key to scaling up a great variety of kernel learning algorithms, with prominent examples including (Williams & Seeger, 2001; Fowlkes et al., 2004; Drineas & Mahoney, 2005; Fine et al., 2001; Achlioptas & McSherry, 2001).

Low-rank matrix decomposition has gained great popularity in tackling large volume of data. However, there are still some concerns with existing approaches. First, most of them are intrinsically unsupervised and only focus on numerical approximation of given ma-trices. When confronted with a specific learning task, however, we believe that incorporating prior knowledge (such as partially labeled samples or grouping constraints) will lead to more effective decomposition and improved performance. Second, many decomposition methods, such as QR decomposition, incomplete Choleskey factorization (Bach & Jordan, 2002), work in a transductive manner. That means the factorization can only be computed for samples available in the training stage. It is not straightforward to generalize the decomposition to new samples, and the difficulty becomes more pronounced if (partial) label information is considered.

To solve the problems, in this paper we propose a novel low-rank decomposition algorithm that incorporates side information in producing desired results. We achieve this by generalizing the Nyström method in a novel way. The Nyström method is a sampling based approach and has gained great popularity in unsupervised kernel low-rank approximation, with both theoretical performance guarantees and empirical successes (Williams & Seeger, 2001; Drineas & Mahoney, 2005; Talwalkar et al., 2008). Our main novelty is to provide an interesting interpretation of the matrix completion view of the Nyström method as a bilateral extrapolation of a dictionary kernel, and generalize it to incorporate prior information in computing improved low-rank decompositions. Our approach has two important advantages. First, it has a flexible, generative structure that allows us to generalize computed low-rank factorizations to arbitrary new samples. Second, both the space and time complexities of our approach are linear in the sample size, rendering great efficiency in learning a useful low-rank kernel.

The rest of the paper is organized as follows. In Section 2, we introduce the Nyström method. In Section 3, we propose the generalized Nyström low-rank decomposition using priors. In Section 4 we discuss related work. In Section 5 we report empirical evaluations. The last section concludes the paper.

## 2. Nyström Method

The Nyström method is a sampling based algorithm for approximating large kernel matrices and their eigen-systems. It originated from solving integral equations and was introduced to the machine learning community by (Williams & Seeger, 2001; Fowlkes et al., 2004; Drineas & Mahoney, 2005).

Given a kernel function $k(\cdot, \cdot)$ and a sample set with underlying distribution $p(\cdot)$, the Nyström method aims at solving the following integral equation

$$\int k(\mathbf{x}, \mathbf{y}) p(\mathbf{y}) \phi_i(\mathbf{y}) d\mathbf{y} = \lambda_i \phi_i(\mathbf{x}).$$

Here $\phi_i(\mathbf{x})$ and $\lambda_i$ are the $i$th eigenfunction and eigenvalue of the operator $k(\cdot, \cdot)$ with regard to $p$. The idea here is to draw a set of $m$ samples $\mathcal{Z}$, called landmark points, from the underlying distribution and approximate the expectation with the empirical average as

$$\frac{1}{m} \sum_{j=1}^{m} k(\mathbf{x}, \mathbf{z}_j) \phi_i(\mathbf{z}_j) = \lambda_i \phi_i(\mathbf{x}) \quad (1)$$

By choosing $\mathbf{x}$ in (1) as $\mathbf{z}_1, \mathbf{z}_2, ..., \mathbf{z}_m$ as well, the following eigenvalue decomposition can be obtained

$$W \phi_i = \lambda_i \phi_i, \quad (2)$$

where $W \in R^{m \times m}$ is the kernel matrix defined on landmark points, $\phi_i \in R^{m \times 1}$ and $\lambda_i$ are the $i$th eigenvector and eigenvalue of $W$.

In practice, given a large data set $\mathcal{X} = \{\mathbf{x}_i\}_{i=1}^{n}$, the Nyström method randomly selects $m$ landmark points $\mathcal{Z}$ with $m \ll n$, and computes the eigenvalue decomposition of $W$. Then the eigenvectors of $W$ are extrapolated to the whole sample set by (1). Interestingly, the Nyström method is shown to implicitly reconstruct the whole $n \times n$ kernel matrix $K$ by

$$K \approx E W^{\dagger} E^{\top}. \quad (3)$$

Here $W^{\dagger}$ is the pseudo-inverse, and $E \in R^{n \times m}$ is the kernel matrix defined on the sample set $\mathcal{X}$ and landmark points $\mathcal{Z}$. The Nyström method requires $O(mn)$ space and $O(m^2 n)$ time, which are linear in the sample size. It has drawn considerable interest in applications such as clustering and manifold learning (Talwalkar et al., 2008) (Zhang & Kwok, 2010), Gaussian processes (Williams & Seeger, 2001), and kernel methods (Fine et al., 2001).

## 3. Generalized Nyström Low-rank Decomposition

### 3.1. Bilateral Extrapolation of Dictionary Kernel

We first present an interesting interpretation of the matrix completion view of the Nyström method (3). It reconstructs $ij$th entry of the kernel matrix as

$$K_{ij} = E_i W^{\dagger} E_j^{\top}, \quad (4)$$

where $E_i \in R^{1 \times m}$ is the $i$th row of the extrapolation matrix $E$. In other words, the similarity between any two samples $\mathbf{x}_i$ and $\mathbf{x}_j$ is constructed by first computing their respective similarities to the landmark set ($E_i$ and $E_j$), and then modulated by the inverse of the similarities among the landmark points, $W^\dagger$. With regards to this we have the following proposition.

**Proposition 1** *Given $m$ landmark points $\mathcal{Z}$, use (4) to construct the similarity between any two samples, $\boldsymbol{x}_i$ and $\boldsymbol{x}_j$. Let $\boldsymbol{z}_p$ and $\boldsymbol{z}_q$ be the closest landmark point to $\boldsymbol{x}_i$ and $\boldsymbol{x}_j$, respectively. Let $d_p = \|\boldsymbol{x}_i - \boldsymbol{z}_p\|$, and $d_q = \|\boldsymbol{x}_j - \boldsymbol{z}_q\|$. Let the kernel function $k(\cdot,\cdot)$ satisfy $k(x,y) - k(x,z) \leq \eta\|y - z\|$, and $c = m \max k(\cdot,\cdot)$. Then the reconstructed similarity $K_{ij}$ and the $pq$th entry of the $W$ will have the following relation*

$$|K_{ij} - W_{pq}| \leq \sqrt{m}\eta(cd_p + cd_q + \sqrt{m}\eta d_p d_q)\|W^\dagger\|_F.$$

**Proof 1** *Let $E_i = [k(\boldsymbol{x}_i,\boldsymbol{z}_1)\ k(\boldsymbol{x}_i,\boldsymbol{z}_2)\ ...k(\boldsymbol{x}_i,\boldsymbol{z}_m)\ ]^\top$, and $W_p = [k(\boldsymbol{z}_p,\boldsymbol{z}_1)\ k(\boldsymbol{z}_p,\boldsymbol{z}_2)\ ...k(\boldsymbol{z}_p,\boldsymbol{z}_m)\ ]^\top$, and define $\Delta_p = E_i - W_p$, $\Delta_q = E_j - W_q$. We have $|\Delta_p|^2 = \sum_{o=1}^m (k(\boldsymbol{x}_i, \boldsymbol{z}_o) - k(\boldsymbol{z}_p, z_o))^2 \leq m\eta^2 d_p^2$, similarly, $|\Delta_q|^2 \leq m\eta^2 d_q^2$. We also have $|E_i|, |E_j| \leq c$, then we have*

$$\begin{aligned}
&|K_{ij} - W_{pq}| \\
=\ & |(W_p + \Delta_p)^\top W^\dagger (W_q + \Delta_q) - W_{pq}| \\
=\ & |(W_p^\top W^\dagger W_q - W_{pq}) - W_p^\top W^\dagger \Delta_q \\
& -\Delta_p^\top W^\dagger W_q + \Delta_p W^\dagger \Delta_q| \\
\leq\ & c(\Delta_p + \Delta_q)\cdot\|W^\dagger\|_F + |\Delta_p|\cdot|\Delta_q|\cdot\|W^\dagger\|_F \\
=\ & (c\Delta_p + c\Delta_q + \Delta_p\Delta_q)\|W^\dagger\|_F \\
=\ & \sqrt{m}\eta(cd_p + cd_q + \sqrt{m}\eta d_p d_q)\|W^\dagger\|_F,
\end{aligned}$$

*Here we used the equality $W_p W^\dagger W_q^\top = W_{pq}$, since $W_p$ and $W_q$ are the $p$th and $q$th row (column) of $W$.*

Proposition 1 gives an interesting interpretation of the kernel reconstruction mechanism of the Nyström method (4). If $\mathbf{x}_i$ and $\mathbf{x}_j$ happen to overlap with a pair of landmark points, $\mathbf{z}_p$ and $\mathbf{z}_q$, then $K_{ij} = W_{pq}$, i.e., the $pq$th entry of $W$ will be extrapolated exactly onto $(\mathbf{x}_i, \mathbf{x}_j)$. In case $\mathbf{x}_i$ and $\mathbf{x}_j$ do not overlap with any landmark point, the difference between $K_{ij}$ and $W_{pq}$, with $\mathbf{z}_p$ and $\mathbf{z}_q$ being the closest landmark points to $\mathbf{x}_i$ and $\mathbf{x}_j$, will be bounded[1] by the distances $\|\mathbf{x}_i - \mathbf{z}_p\|$ and $\|\mathbf{x}_j - \mathbf{z}_q\|$. The smaller the distances, the closer $K_{ij}$ and $W_{pq}$. In other words, the similarity matrix $W$ on the landmark points serves as a *dictionary kernel*, whose entries are extrapolated bilaterally onto any pairs of samples $(\mathbf{x}_i, \mathbf{x}_j)$ according to the proximity relation between landmark points and samples, and the reconstruction is exact on the landmark points $\mathcal{Z}$ which serve as the "nodes" for extrapolation.

---

[1] A tighter bound is still open and being investigated.

### 3.2. Including Side Information

The kernel extrapolation view of the Nyström method (Proposition 1) inspires us to generalize it to handle prior constraint in learning a low-rank kernel. Note that quality of the dictionary will have a large impact on the whole kernel matrix. In the original Nyström method (3), the dictionary kernel $W$ is simply computed as the pairwise similarity between landmark points, which can deviate from an "ideal" one. Therefore, instead of using such an "unsupervised" dictionary, we propose to learn a new dictionary kernel that better coincides with given side information.

Suppose we are given a set of labeled and unlabeled samples[2]. Let $\mathcal{Z}$ be a set of $m$ pre-selected landmark points. Let $E \in R^{n \times m}$ be the extrapolation kernel matrix between samples $\mathcal{X}$ and landmark $\mathcal{Z}$, and let $E_l \in R^{l \times m}$ be the rows of $E$ corresponding to labeled samples. For simplicity, let $S_0 = W^\dagger$ denote the inverse of the dictionary kernel in the standard Nyström method (3). Our task is to learn (the inverse of) a new dictionary kernel, denoted by $S$, subject to the following considerations:

1. **unsupervised information**: the reconstructed kernel $ESE^\top$ should preserve the structure of the original kernel matrix $K$, since $K$ encodes important pairwise relation between samples.

2. **supervised information**: the reconstructed kernel on the labeled samples, $E_l S E_l^\top$, should be consistent with the given side information.

To achieve the first goal, note that in the standard Nyström method, $EW^\dagger E^\top$ provides an effective approximation of $K$. Therefore, we use $S_0 = W^\dagger$ as a prior for the (inverse) dictionary kernel $S$, namely, they should be close under some distance measurement. To achieve the second goal, we use the concept of kernel target alignment (Cristianini et al., 2002) and require that the reconstructed kernel, $E_l S E_l^\top$, is close to the ideal kernel $K_l^*$ defined on labeled samples. The ideal kernel is defined as (Kwok & Tsang, 2003)

$$[K_l^*]_{ij} = \begin{cases} 1 & \mathbf{x}_i, \mathbf{x}_j \text{ in the same class} \\ 0 & \text{otherwise}. \end{cases} \quad (5)$$

We therefore arrive at the following problem

$$\min_{S \in R^{m \times m}} \lambda\|S - S_0\|_F^2 + \|E_l S E_l^\top - K_l^*\|_F^2 \quad (6)$$

$$s.t.\ S \succeq 0.$$

---

[2] In case side information is in the form of grouping constraints, discussion is in Section 3.3.

Here, we used the Euclidian distance to measure the closeness between two matrices. Note that in (Cristianini et al., 2002), the closeness between two kernel matrices is measured by their inner product $\langle K, K' \rangle = \sum_{i,j} K_{ij} K'_{ij}$. Since $\|K - K'\|_F^2 = \langle K, K \rangle + \langle K', K' \rangle - 2\langle K, K' \rangle$, minimizing the Euclidian distance is related to maximizing the alignment. We choose the Euclidian distance here because we can then use the normalized kernel alignment score afterwards as an independent measure to choose the hyper-parameter $\lambda$. Details will be discussed in Section 3.8.

We call our method generalized Nyström low-rank decomposition, which has several desirable properties. First, as long as the inverse dictionary kernel $S$ is psd, the resultant kernel $ESE^\top$ will also be psd; second, the rank of the kernel matrix can be easily controlled by the landmark size; this can be computationally much more efficient than learning a full kernel matrix subject to rank constraint; third, the extrapolation (4) is "generative" and allows us to compute the similarity between any pair of samples; this means the learned kernel matrix generalizes easily to new samples. Since the dictionary kernel is learned with prior information, the generalization to new samples naturally incorporates such information, which provides much convenience in updated environments.

### 3.3. Side Information as Grouping Constraints

Given a set of grouping constraints (must-link and cannot-link pairs), denoted by $\mathcal{I}$. Let $\mathcal{X}_I$ be the subset of samples with such constraint. Then we define $\mathbf{T} \in R^{|\mathcal{X}_I| \times |\mathcal{X}_I|}$ such that

$$\mathbf{T}_{ij} = \begin{cases} 1 & (\mathbf{x}_i, \mathbf{x}_j) \in \mathcal{X}_I \\ 0 & otherwise. \end{cases}$$

Then our objective can be written conveniently as

$$\min_{S \in R^{m \times m}} \lambda \|S - S_0\|_F^2 + \|\mathbf{T} \odot (E_I S E_I^\top) - K_I^*\|_F^2$$
$$s.t. \quad S \succeq 0.$$

Here $K_I^*$ is defined similarly as in (5).

### 3.4. Optimization

The objective (6) is convex regard to $S$, and the psd constraint $S \succeq 0$ is also convex. Therefore (6) is a smooth convex problem with a global optimum.

Note that $S$ is a matrix with only $m^2$ variables, where $m \ll n$ is a user defined value. Therefore the problem (6) involves only light optimization load. We use the gradient mapping strategy (Nemirovski, 1994) that is composed of iterative gradient descent equipped with a projection step to find the optimal solution. Given an initial solution $S^{(t)}$, we update it by

$$S^{(t+1)} = S^{(t)} + \eta^{(t)} \nabla_{S^{(t)}}, \tag{7}$$

where $\nabla_S$ is the gradient of the objective $J$ (6) at $S$,

$$\nabla_S = 2\lambda(S - S_0) + 2E_l^\top (E_l S E_l^\top - K_l^*) E_l.$$

The step length $\eta^{(t)}$ is determined by the Armijo-Goldstein rule (Nemirovski, 1994). In particular, we start from an initial, small scalar $A$, and solve the following problem

$$B_A^* = \arg\min_{B \succeq 0} tr(\nabla_{S^{(t)}} B) + \frac{A}{2} \|B - S^{(t)}\|_F^2. \tag{8}$$

This is a standard matrix nearness problem with psd constraint, and $B_A^*$ can be computed in closed form as $S^{(t)} - \frac{1}{A} \nabla_{S^{(t)}}$ removed of negative eigenvectors/values. Then we examine

$$J(B_A^*) \leq J(S^{(t)}) + tr\left(\nabla_{S^{(t)}} (B_A^* - S^{(t)})\right) + \frac{A}{2} \|B_A^* - S^{(t)}\|_F^2.$$

If this inequality is violated, then we increase $A$ by a constant times and re-calculate (8) until the relation holds. Then we use $\eta^{(t)} = \frac{1}{A}$ as the step length for (7).

After the descent step, we project the iterate $S^{(t+1)}$ onto the set of positive semi-definite cones as follows

$$S^{(t+1)} \leftarrow U^{(t+1)} \Lambda_+^{(t+1)} (U^{(t+1)})^\top,$$

where $U^{(t+1)}$ and $\Lambda^{(t+1)}$ are the eigenvectors and eigenvalues of $S^{(t+1)}$ (7), and

$$\Lambda_+^{(t+1)}{}_{ii} = \begin{cases} \Lambda_{ii}^{(t+1)} & if \ \Lambda_{ii}^{(t+1)} \geq 0; \\ 0 & otherwise. \end{cases} \tag{9}$$

One can also use more advanced approaches such as the Nesterov's method (Nemirovski, 1994) to improve the convergence rate. We do not explore details here because the size of our optimization problem is small and empirically it converges quickly due to a principled initialization (see next subsection).

### 3.5. Initialization

In this section, we propose a closed-form initialization which helps us quickly locate the optimal solution. The basic idea is to drop the psd constraint in (6) and compute the vanishing point of the gradient, i.e., $\frac{\partial J(S)}{\partial S} = 0$, which leads to

$$\lambda S + E_l^\top E_l S E_l^\top E_l = E_l^\top K_l^* E_l + \lambda S_0.$$

Then we have

$$S + PSP^\top = Q, \quad (10)$$
$$\text{where} \quad P = \frac{1}{\sqrt{\lambda}}(E_l^\top E_l),$$
$$Q = S_0 + \frac{1}{\lambda} E_l^\top K_l^* E_l.$$

Equation (10) can be solved as follows. Suppose the diagonalization of $P$ is $P = U\Lambda U^\top$, and define $S = U\tilde{S}U'$, $Q = U\tilde{Q}U^\top$, then it can be written as

$$U\tilde{S}U^\top + U\Lambda\tilde{S}\Lambda U^\top = U\tilde{Q}U^\top \;\rightarrow\; \tilde{S} + \Lambda\tilde{S}\Lambda^\top = \tilde{Q}.$$

Since $\Lambda$ is diagonal, this becomes $m^2$ equations

$$\tilde{S}_{ij} + \Lambda_{ii}\Lambda_{jj}\tilde{S}_{ij} = Q_{ij}, \quad 1 \le i, j \le m.$$

Therefore we have a closed form solution of $S$, as

$$S = U\tilde{S}U^\top, \; \text{where} \; [\tilde{S}]_{ij} = \frac{\tilde{Q}_{ij}}{1 + \Lambda_{ii}\Lambda_{jj}}.$$

After computing $S$, we then project it onto the set of positive semi-definite cones similar to (9). Such an initial solution can be deemed as the closest psd matrix to the unconstrained version of (6). Empirically, such an initial solution alone already leads to satisfactory prediction result.

### 3.6. Landmark Selection

Selection of the landmark points $\mathcal{Z}$ in the Nyström method can greatly affect its performance. Preferably, landmark points should allow faithful reconstruction of the global similarity landscape. We used the $k$-means based sampling scheme (Zhang & Kwok, 2010) which has shown to consistently outperform other popular landmark selection schemes such as random sampling.

### 3.7. Complexities

The space complexity of our algorithm is $O(mn)$, where $n$ is sample size and $m$ the number of landmark points. Computationally, it only requires repeated eigenvalue decomposition of $m \times m$ matrices, and a single multiplication between the $n \times m$ extrapolation $E$ and the $m \times m$ dictionary kernel $S$. The over all complexity is $O(m^2 n) + O(t \log(\mu_{\max})m^3)$, where $t$ is the number of gradient mapping iterations, and $\mu_{\max}$ is the maximum eigenvalue of the Hessian. This is because $A$ (8) is bounded by $\mu_{\max}$ and one can always find a suited step-length in $\log(\mu_{\max})$ steps. Empirically, with the initialization in Section 3.5, only a few iterations is needed. Therefore $t$ is a small integer and our algorithm has a linear time and space complexity.

### 3.8. Selecting $\lambda$

The hyper-parameter $\lambda$ in (6) can be difficult to choose if the side information (e.g., partially labeled samples) is limited. Here we propose a heuristic to choose $\lambda$. Note that the objective (6) contains two residuals, $S_0 - S$, and $E_l S E_l^\top - K_l^*$, in terms of the Euclidian distance, which are additive and requires a tradeoff parameter $\lambda$. Here, we use a new criterion with certain invariance property to re-evaluate the goodness of fit of the solution. More specifically, we used normalized kernel alignment (NKA) (Cortes et al., 2010) between kernel matrices,

$$\rho[K, K'] = \frac{\langle K_c {K'_c}^\top \rangle_F}{\|K_c\|_F \|K'_c\|_F}, \quad (11)$$

where $K_c$ is double-centralized $K$. The NKA score always has a magnitude that is smaller than 1. It is independent of the scale of the solution, and is multiplicative by nature. Let $S(\lambda)$ be the optimum of (6) for a fixed $\lambda$. Then we choose the best $\lambda$ as follows

$$\lambda^* = \arg\max_{\lambda \in \mathcal{G}} \rho\left[S(\lambda), S_0\right] \cdot \rho\left[E_l S(\lambda) E_l^\top, K_l^*\right]. \quad (12)$$

Here $\mathcal{G}$ is the set of candidate $\lambda$'s. The criterion (12) has the following properties: (1) it is scale invariant, and does not require any extra trade-off parameter due to its multiplicative form; (2) the first term measures the closeness between $S$ and $S_0$, related to unsupervised structures of kernel matrix; the second term is on the closeness between $E_l S E_l^\top$ and $K_l^*$, related to side information; therefore the criterion faithfully reflects what (6) optimizes but on the other hand is numerically different; (3) a higher alignment (second term in (12)) indicates existence of a good predictor with higher probability (Cortes et al., 2010); (4) computation of the criterion does not require any extra validation set, which is suited if only limited training samples are available. Therefore, this is an informative criterion to measure the quality of solution. Empirically, it correlates nicely with the prediction accuracy on the test samples, as will be reported in Section 5.

## 4. Related Work

This section discusses several lines of work on using side information in low-rank kernel matrices.

One is to rectify standard numerical low-rank decomposition procedures by injecting supervised information. An excellent example is the Choleskey with Side Information (SCI) (Bach & Jordan, 2005). The algorithm is iterative and in each step, the column of the kernel matrix that maximally reduces the hybrid of the matrix approximation error and a linear prediction error is selected. One difficulty with the greedy scheme is that the approximation error takes $O(n^2)$ time to compute, and an upper bound has to be used instead, which may adversely affect the result. The algorithm takes into account the label information and can reduce the rank of factorization needed in a kernel classifier. Our approach was motivated similarly but has important differences. First, the CSI method assumes that labels of all training instances are given (extension to semi-supervised setting will require non-trivial modifications of the algorithm); in comparison, we consider the more generalized semi-supervised learning scenario. Second, the procedure is transductive and there seems to be lacking principled ways to compute factorizations for new samples; wheares our approach generalize easily to new samples by design.

The second line is low-rank kernel learning. Although kernel learning has drawn considerable interest, algorithms on learning low-rank kernel matrices are not very abundant (Kulis et al., 2009), in particular those in a computationally efficient way. Lanckriet et al. studied transductive kernel learning through a general, semi-definite programming (SDP) framework. The rank of the learned kernel can be controlled by choosing kernel matrix as a convex combination of a small number of base kernels. However, even special cases of it (QCQP) are still computationally expensive, with at least cubical time complexity in sample size. Kulis et al. proposed to learn a low-rank kernel by minimizing its divergence with an initial low-rank base kernel subject to distance/similarity constraints. They applied the Bregman divergence which naturally preserves the low-rankness and positive semi-definiteness of solution. The algorithm improves in efficiency, but in general it still has quadratic space and time complexities with the sample size. In (Shalit et al., 2010) an online learning algorithm is proposed on the manifold of low-rank matrices, which consists of iterative gradient step and second-order retraction. In (Machart et al., 2011), a novel low-rank kernel learning approach was proposed for regression via the use of conical combinations of base kernels and a stochastic optimization framework. Again, most of these algorithms are transductive and how to generalize the learned kernel to new samples still remains open.

The third line involves spectral kernel learning, which builds a kernel matrix using eigenvectors and rectified eigenvalues of the graph Laplacian. Transformation of the eigen-spectrum can be achieved analytically, such as in (Kondor & Lafferty, 2007) (Chapelle et al., 2003). In (Cristianini et al., 2002) (Cortes et al., 2010), a nonparametric transform is computed by maximizing the alignment with the target. In (Zhu et al., 2004), an extra order constraint on the weight of eigenvectors was adopted. Due to the need to compute kernel eigenvalues, spectral kernel learning requires at least quadratic space and time complexities, or even higher if advanced optimization such as QCQP is involved (Zhu et al., 2004).

## 5. Experiments

This section compares 7 algorithms on learning low-rank kernel: (1) Nyström: standard Nyström method; (2) CSI: Choleskey with Side Information (Bach & Jordan, 2005); (3) Cluster: cluster kernel (Chapelle et al., 2003); (4) Spectral: non-parametric spectral graph kernel (Zhu et al., 2004); (5) TSK: two stage kernel learning algorithm (Cortes et al., 2010); (6) Breg: low-rank kernel learning with Bregman divergence (Kulis et al., 2009); (7) Our method. Most algorithms can learn the $n \times n$ low-rank kernel[3] matrix on labeled and unlabeled samples[4] in the form of $K = GG^\top$, which is fed into SVM for classification. The resultant problem will be a linear SVM using $G$ as training/testing samples (Zhang et al., 2012).

We use benchmark data sets from the SSL data set (Chapelle et al., 2001) and the libsvm data. For each data set, we randomly pick 100 labeled samples evenly among all classes, repeat 30 times and report the averaged classification error on unlabeled data. We used the Gaussian kernel $K(\mathbf{x}_1, \mathbf{x}_2) = \exp(-\|\mathbf{x}_1 - \mathbf{x}_2\|^2/b)$. Parameter selection is difficult in semi-supervised learning, so, we choose the kernel width as the averaged pairwise squared distances between samples. Empirically, this gives reasonable performance compared with the best kernel width from some pre-defined candidates. For method (5) the base kernel are chosen from a set of RBF kernels whose widths are factors of the averaged pairwise distance as in (Cortes et al., 2010). For the regularization parameter $C$ in linear SVM, we use the heuristic implemented in liblinear package (Fan et al., 2008). Most codes are in matlab (for method (2) we used codes in (Bach & Jordan, 2005) with core functions in C) and run on a PC with 2G memory and 2.8GHz processor.

Results are reported in Table 1. Methods statistically better than others with a confidence level that is at least 95% (paired $t$-test) are highlighted. Note that method (1) does not use label information nor unla-beled data in training, therefore as a baseline method it is very efficient. Method (2) is very efficient because it only uses labeled samples for training (with C implementation). Method (3), (4), (5) require eigenvalue decomposition of the kernel matrix (or graph Lapacian), therefore they are computationally more expensive. Method (6) may require many iterations to converge. Our approach is very efficient and can be orders of magnitudes faster than some other methods.

---

[3] The rank of the learned kernel is set to be 10% of sample size (or a fixed number if sample is too large).

[4] Method (2) uses some heuristics to compute the kernel matrix between labeled and unlabeled samples, since only labeled samples are used in training.

Table 1. Performance of different methods; top row: mean/std of error (%); bottom row: average time (in seconds).

| data size/dim/#classes | Nyström | CSI | Spectral | Cluster | TSK | Breg | Ours |
|---|---|---|---|---|---|---|---|
| g241C 1500/241/2 | 27.09±2.16 0.8 | 22.37±1.80 0.8 | 23.64±1.28 108.2 | 26.59±3.96 63.3 | 23.47±1.09 2.7 | 26.31±1.81 17.7 | **21.57±0.85** 1.2 |
| Digit1 1500 | 5.67±1.25 0.8 | 4.97±0.84 0.8 | **3.87±1.71** 70.5 | 5.53±1.01 0.5 | 6.86±1.25 2.8 | 6.01±1.62 27.3 | 4.71±0.71 1.2 |
| USPS 1500/241/2 | 10.67±3.46 0.7 | 8.60±2.55 0.7 | 11.28±0.51 78.4 | **6.96±1.14** 149.1 | 8.64±2.64 2.5 | 11.45±2.34 27.5 | 8.66±1.14 1.0 |
| coil 1500/241/6 | **19.04±2.90** 0.8 | 19.46±2.88 0.8 | 31.21±10.06 31.0 | 19.38±2.40 41.7 | 22.04±3.94 4.3 | 19.57±3.41 19.8 | 19.60±3.20 1.2 |
| coil2 1500/241/2 | 12.98±4.60 0.8 | 12.44±4.68 0.6 | 14.57±3.28 39.1 | 13.61±2.24 47.6 | **11.76±2.65** 2.4 | 14.58±3.17 27.3 | 11.83±3.40 1.1 |
| Text 1500/11960/2 | 27.10±2.22 37.9 | 22.69±1.86 4.9 | 24.90±2.04 40.5 | 27.8±2.19 287.9 | 23.29±1.61 27.6 | 23.66±0.90 44.3 | **22.10±1.32** 28.2 |
| german 1000/24/2 | 40.31±2.74 0.2 | 37.73±2.27 0.6 | 33.83±10.51 41.6 | **31.90±3.42** 2.5 | 33.93±3.20 1.2 | 38.42±2.95 512.3 | 36.84±3.37 0.4 |
| usps49 1296/256/2 | 2.73±0.72 0.7 | 2.36±1.04 0.8 | **1.58±0.40** 26.6 | 1.74±0.23 8.4 | 1.82±0.73 2.5 | 3.04±0.39 18.5 | 1.67±0.34 1.0 |
| usps27 1367/256/2 | 1.25±0.27 0.6 | 1.29±0.27 0.6 | 1.98±0.33 7.82 | 1.25±0.31 14.6 | 1.24±0.21 3.1 | 1.52±0.47 22.5 | **1.10±0.30** 1.0 |
| adult1a 1605/123/2 | 29.33±2.96 0.6 | 25.69±2.30 0.7 | 27.27±1.99 32.4 | 24.50±2.68 14.1 | 26.62±3.57 2.7 | 32.62±2.22 17.5 | **23.93±2.06** 1.2 |
| dna 2000/180/3 | 15.92±2.03 1.8 | 15.45±1.36 1.2 | 15.68±1.51 62.2 | 20.87±2.16 48.5 | 18.15±1.94 25.1 | 15.80±0.16 38.1 | **15.50±1.83** 2.4 |
| segment 2310/29/7 | 9.60±1.49 1.7 | **9.39±1.14** 2.1 | 15.51±3.10 83.2 | 17.91±2.82 19.9 | 9.50±1.32 8.8 | 10.08±1.79 22.6 | 9.59±1.20 2.7 |
| svmgd1a 3089/4/2 | 5.22±0.97 0.6 | 6.55±0.33 4.5 | 6.54±0.91 205.5 | 5.18±1.79 47.6 | 4.80±0.84 8.1 | 5.51±1.45 17.6 | **4.40±0.83** 1.5 |
| satimage 6435/36/6 | 18.70±1.82 1.7 | 18.54±0.97 2.2 | 19.39±1.63 285.5 | 20.78±2.36 197.7 | **17.13±0.86** 3090.0 | 18.52±1.82 638.4 | 17.88±1.40 2.8 |
| usps-full 7291/256/10 | 14.47±1.43 5.5 | **13.43±1.51** 3.8 | 14.32±1.81 521.3 | 14.79±2.39 363.9 | 13.93±1.59 4163.0 | 14.25±2.29 1418.5 | 13.68±1.42 6.1 |
| mnist 70000/780/10 | 25.12±1.85 80.4 | 24.70±1.26 33.3 | - - | - - | - - | 23.96±1.73 151.7 | **21.85±1.77** 82.3 |

From table 1, we can see that on most data sets, algorithms using labels in kernel learning outperform the baseline algorithm (method 1), indicating the value of side information. Our approach is competitive with stat-of-the-art kernel learning algorithms. On the largest data set mnist, method (3), (4), (5) can not run on our PC due to the huge memory consumptions; in comparison, our approach is very efficient and gives the lowest error rate on this data set. We also examine the alignment score (12) used to choose the hyperparameter $\lambda$ in Figure 1. As can be seen, the score correlates nicely with the classification accuracy.

## 6. Conclusions

In this paper, we proposed an efficient kernel low-rank decomposition algorithm endowed with a flexible, non-parametric reconstruction mechanism, while being capable of handling side information. It shows significant performance gains in benchmark learning tasks. In the future, we will couple the dictionary learning with specific classifier such as an SVM to further improve the prediction performance. Another interesting direction is the learning of a *sparse* dictionary and its application in information retrieval problems.

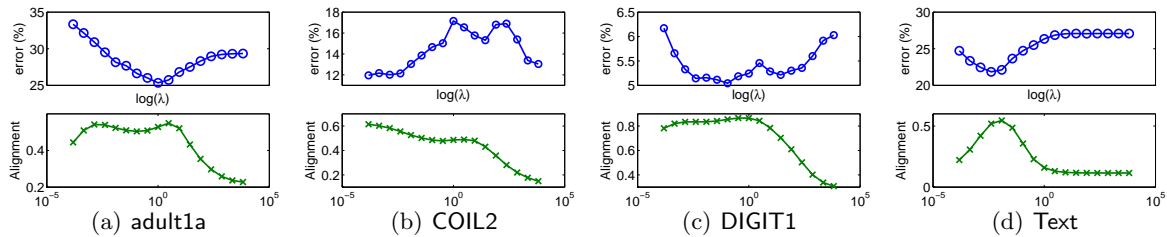

*Figure 1.* The alignment score defined in (12) (bottom box) correlates with the classification error (top box). By choosing the $\lambda$ that gives the highest alignment, we can obtain a satisfactory prediction.